%% file: driss.tex
\documentclass[runningheads]{llncs}
\usepackage{graphicx}
\usepackage[export]{adjustbox}
\usepackage{algorithm}
\usepackage{placeins}

\usepackage[caption=false]{subfig}
\usepackage{booktabs}       
\usepackage{array,multirow,makecell}
\usepackage{rotating}

\usepackage[dvipsnames]{xcolor}

\usepackage{xspace}
\def\ie{{\em i.e.,}\xspace}

\graphicspath{ {./figures/} }
\begin{document}
\title{Deep Reinforcement Learning for $5 \times 5$ Multiplayer Go}
\titlerunning{Deep Reinforcement Learning for $5 \times 5$ Multiplayer Go}
%
\author{ Brahim Driss\inst{1} \and Jérôme Arjonilla\inst{1} \and Hui Wang\inst{1} \and Abdallah Saffidine\inst{2} \and Tristan Cazenave\inst{1}
}


%
\authorrunning{Driss et al.} 
%
\institute{LAMSADE, Université Paris Dauphine - PSL, CNRS, Paris, France \email{Tristan.Cazenave@dauphine.psl.eu} \email{brahim.driss0@dauphine.eu} \email{jerome.arjonilla@hotmail.fr} \email{hui.wang@dauphine.psl.eu} \and
University of New South Wales, Sydney, Australia \\      \email{abdallah.saffidine@gmail.com}}


\maketitle              
\begin{abstract}
In recent years, much progress has been made in computer Go and most of the results have been obtained thanks to search algorithms (Monte Carlo Tree Search) and Deep Reinforcement Learning (DRL). In this paper, we propose to use and analyze the latest algorithms that use search and DRL (AlphaZero and Descent algorithms) to automatically learn to play an extended version of the game of Go with more than two players. We show that using search and DRL we were able to improve the level of play, even though there are more than two players. 
\end{abstract}

\input{body}

\bibliographystyle{splncs04}
\bibliography{references}

\end{document}

%% file: body.tex
\section{Introduction}


Due to a huge game tree complexity, the game of Go has been an important source of work in the perfect information setting. In 2007, search algorithms have been able to increase drastically the performance of computer Go programs \cite{Coulom2006,Coulom2007,Kocsis2006,Gelly2011AI}. In 2016, AlphaGo has been able to beat a strong professional player for the first time \cite{Silver2016MasteringTG}. This great success has been achievable thanks to a combination of two key elements: Search (Monte Carlo Tree Search \cite{browne2012survey}) and Learning (Reinforcement Learning) methods \cite{Silver2016MasteringTG,silver2017mastering,silver2018general}. Currently, the level of play of such algorithms is far superior to those of any human player.


Even though, the game of Go has been given great interest, less has been done on variants of the game. In practice, there exist many variants of the game of Go such as Blind Go \cite{chou:inria-00625849} (where the players cannot see the board), Phantom Go \cite{cazenave2005phantom} (where the players cannot see the opponent stones) or Capture Go (where the game is finished when the first player to capture a stone wins). In this paper, we study the variant Multiplayer Go. As the name suggests, Multiplayer Go is a variant of the game of Go where there are more than two players. Going from two-player Go to Multiplayer Go makes the game even more complex. 


In this paper, we propose to apply and analyze the latest developments in the game of Go to the game of Multiplayer Go. More specifically, we are using search and reinforcement learning method such as AlphaZero \cite{Silver2016MasteringTG} and Descent \cite{CohenSolal2020LearningTP,cohen2020minimax} for the game of Multiplayer Go. In past work, \cite{Cazenave08multi}
has described multiple UCT algorithms with different multi-agent behaviors (coalitions, paranoid or with alliance) for the game of Multiplayer Go and in \cite{Cazenave15}, they successfully improve past performances using GRAVE, a heuristic method for MCTS algorithms. An adaptation of AlphaZero to multiplayer games was also used in \cite{Petosa2019MultiplayerA} for the multiplayer versions of Tic-Tac-Toe and Connect 4.

The paper is organized as follows: the second section presents the game of Multiplayer Go, section three presents the algorithms we have been using to analyze the game, section four presents our results and the last section summarizes our work and future work.

\section{Multiplayer Go}



The game of Go is a strategic board game with perfect information, played by two players. Each player aims at capturing more territory than their opponent by placing stones on the board. One is playing black stones and the second is playing white stones. At each turn, one player is acting and placing a stone on a vacant intersection of the board. After being placed, a stone cannot be moved or removed by the player. Nevertheless, a player stone can be removed by its opponent if the latter successfully surrounds the stone on all orthogonally adjacent points. The game ends when no player is able to make a move or until none of them wishes to move. 

For the scoring, there exist multiple rules, in our case, we have used the Chinese rule \ie the winner of the game is defined by the number of stone that a player has on the board, plus the number of empty intersections surrounded by that player's stones and komi (bonus added to the second player as a compensation for playing second).

\begin{figure*}
		\centering
		\includegraphics[scale=0.4]{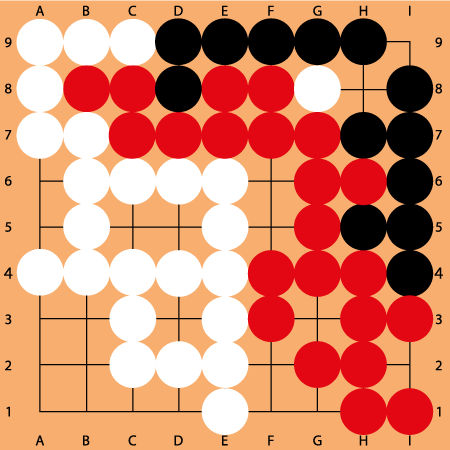}
		\caption{A game of Multiplayer Go.}
		\label{fig:go_game}
\end{figure*}

Multiplayer Go is a variant of the game of Go with more than two players. In our case, we have added a third player which is playing a third color, red. An example is provided in Figure \ref{fig:go_game}. By adding a third player, one must be wary to not create a queer game \cite{loeb1996stable}. 

A queer game is when, in some positions, no player can force a win. As an example, in some positions, even if a player is sure to lose, they can still have an impact on the winner by adapting this strategy. As a consequence, coalitions can arise in order to help defeat another opponent. 

As a preliminary matter, we have compared the performance of two different type of rewards, one that uses the winning player as an objective and another one that tries to maximize the score, using Chinese rules. Maximizing the score allows us to reduce the problem of queer games. This analysis had already been carried out in \cite{Cazenave08multi}. For the remainder of the paper, we thus use the objective for each player to maximize his score using Chinese rules.

Even tough the rules are relatively simple, the game of Go is known as an extremely complex one in comparison to other board game such as Chess. With a larger board (common board have a size of $19 \times 19$), the number of legal board positions has been estimated to be $2.1 \times 10^{170}$. Moreover, with a large number of possible actions and longer games on average, the complexity is much greater than that. Worse than that, in our case, the addition of a three-player game adds a significant layer to the complexity of the game. As a consequence, in an effort to study the learning in a multiplayer setup rather than the difficulty of the game, we studied a simplified version of Go on a $5 \times 5$ board. Nevertheless, even being smaller, the version of the game is very complex.


\section{Deep Reinforcement Learning}

In this section, we described the algorithms used for addressing Multiplayer Go. The subsection \ref{subsec:UCT} presents Monte Carlo Tree Search (MCTS) and its variant UCT, the subsection \ref{subsec:AZ} presents AlphaZero and the subsection \ref{subsec:Descent} presents Descent. All the hyperparameters used are explained and defined in Table \ref{tab:training_param} and in Table \ref{networkhyperparams}.

\subsection{Monte Carlo Tree Search}
\label{subsec:UCT}

In our experimentation, we have been using UCT (Upper Confidence bounds applied to Trees), which is a variant of MCTS, as a baseline. Before explaining UCT, we must explain MCTS. 
Monte Carlo Tree Search ~\cite{browne2012survey} is the state of the art in perfect information games. MCTS is a tree search algorithm which works as follows (\romannumeral 1) \textbf{selection} --- select a path of node based on the exploitation policy (\romannumeral 2) \textbf{expansion} --- expand the tree by adding a new child node (\romannumeral 3) \textbf{playout} --- estimate the child node by using an exploration policy (\romannumeral 4) \textbf{backpropagation} --- backpropagate the result obtained from the playout through the nodes chosen during the selection phase.
\\\\
UCT is a variant of MCTS where the selection phase is decided by UCB (Upper Confidence Bounds), a bandit algorithm and where the playout use a random policy. The UCB formula is decomposed on two parts, the first part represents the exploitation \ie it attempts to play the best action observed so far, and the second part represents the exploration \ie it attempts to play an action less visited. 

The formula is defined as follow  :
    \begin{equation}
    UCT(s,a) = Q(s,a) + c\sqrt{\frac{ln[N(s)]}{N(s,a)}}
    \label{ucborigin}
    \end{equation}
where the best action is the one that maximizes the upper confidence bound $UCT(s,a)$, $s$ denotes the state of the game, $a$ is an action possible from the set $A(s)$ which represents all the actions possibles in the state $s$. $Q$ represents the value when playing the action $a$ in the state $s$, $N(s,a)$ is the number of times that the action $a$ has been visited in the state $s$, $N(s)$ represents the number of times that the state $s$ has been visited and $c$ is a variable that help controlling the exploration. 
\\\\
Furthermore, as we are in a multiplayer context, we must use Multiplayer UCT \cite{multi_uct} which is the same algorithm as UCT where the only difference being the score representation. In a multiplayer setup, we get an array instead of a single value, containing the results of the different players.
%
%
%
\\\\
In our experimentation, we use the following hyper-parameters (\romannumeral 1) $n=180$ the number of rollout \ie the number of times the playout is repeated in order to obtain a better approximation of the child node (\romannumeral 2) $c =0.8$.

\subsection{AlphaZero} 
\label{subsec:AZ}

As a famous deep reinforcement learning paradigm, combining online Monte Carlo Tree Search (MCTS) and offline neural network has been widely applied to solve game-related problems, especially known as AlphaGo series programs~\cite{Silver2016MasteringTG,silver2018general,silver2017mastering}. MCTS is used to enhance the policy and the neural network provides the state estimation. 

The neural network based MCTS employs PUCT formula to balance the exploration and exploitation as following :
    \begin{equation}
    PUCT(s,a) =  Q(s,a) + c_{puct}.P_{\theta}(s,a)\frac{\sqrt{N(s)}}{1+N(s,a)}
    \label{pucb}
    \end{equation}
where the best action is the one that maximizes the upper confidence bound $PUCT(s,a)$, $s$ denotes the state of the game, $a$ is an action possible from the set $A(s)$ which represents all the actions possibles in the state $s$. $Q$ represents the value when playing the action $a$ in the state $s$, $N(s,a)$ is the number of times that the action $a$ has been visited in the state $s$, $N(s)$ represents the number of times that the state $s$ has been visited, $c_{puct}$ is a variable that help controlling the exploration and $P_{\theta}(s,a)$ is the estimation of taking action $a$ in the state $s$ according to the policy of the network $\theta$.


The network architecture is similar to the original AlphaZero one, having the board as input and producing two outputs : a probability distribution over moves (policy head) and a vector of score prediction for every player (value head).


\begin{figure*}
	\centering
	\includegraphics[width=\textwidth]{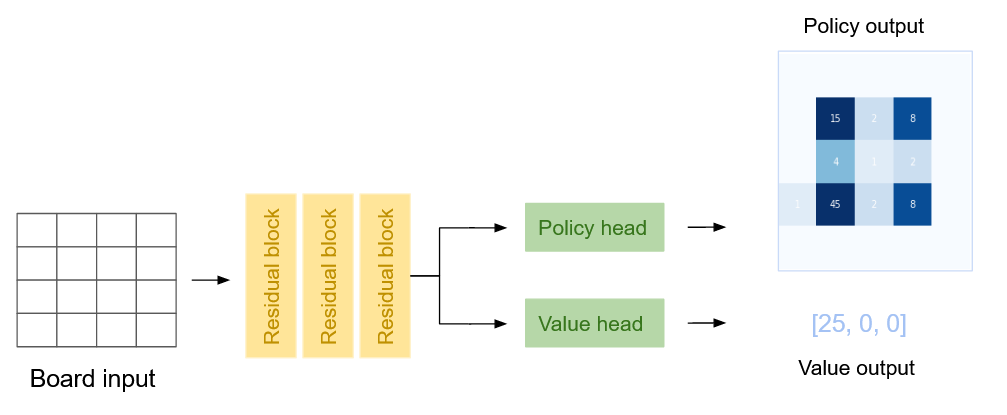}
	\caption{AlphaZero network architecture}
	\label{aznetwork}
\end{figure*}

\subsubsection{Network architecture}


The general architecture of AlphaZero can be found in Figure \ref{aznetwork}. All hyperparameters and differences can be found in Table \ref{networkhyperparams}.

In order to encode the board, we are using a $5 \times 5 $ matrix. The input of the state is represented by 6 channels where each channel is a $5 \times 5$ matrix. The first 3 channels represent the position of one player's stones on the board, and the last 3 channels represent the current player who has to play.

After the encode of the game, 8 residual block of width 128 are placed one after the other. The first residual block take the game encoded as an input.  Each residual block are using convolution kernel of size $3$ with the activation function being ReLU and as we are using a residual block, the input of the layer is also for the next residual block. 


The policy head give the probability of playing each action for the current player \ie for the 25 actions, the policy head return a value between 0 and 1. Thus, the policy head is composed of a 1x1 convolution, outputting a policy distribution in logits. 

The value head outputs 3 values as described in multiplayer UCT \cite{multi_uct} which estimated the value that each player will obtain \ie for the 3 players, the value head returns a score between 0 to 25. Thus, the value head is a fully connected with 3 parallel hidden layers, each one connected to an output layer of size 26 (possible integer scores for each player from 0 to 25 points). 


Usually, in the game of Go, a player has the possibility of passing. However, when tested, and even with the goal of maximizing the score, the agent starts passing even in positions where it is still possible to continue gaining more points. In order to fix this performance issue, we removed the pass from the network, and only allowed passing when no more moves are possible (or only moves that fill the eyes, where an eye is a single empty space inside a group). 


\subsubsection{Warm-Start self-play}

For training our network, we have been using self-play \ie we compete against ourselves, save the data collected and train using this data. AlphaZero self-play starts with randomly initialized networks for the value and policy. Both of them are used, combined with tree search, in order to generate games which are used to improve the networks, leading to better decisions by learning MCTS selected moves (policy improvement) and better value estimation, having access to the games results at the end (value improvement).

Playing moves based on a random policy network will generate games where decisions are almost random, leading to longer training time before observable improvement. In order to accelerate this process, we add UCT agents during self-play, replacing (one or many) AlphaZero agents randomly with a decreasing probability $\epsilon$, where $\epsilon = max( (0.5 - \frac{iteration\_number}{ n\_updates}),0.05 )$.

As a result, in earlier iterations, every AlphaZero agent (each color) has a 50\% probability in every game to be replaced by a UCT agent, this probability decreases every iteration, reaching 5\% by iteration 50 where it stops decreasing.




\begin{table}
\centering
\caption{Training cycle hyperparameters}

\begin{tabular}{l|l|l}
\toprule
\textbf{Hyperparameter} & \textbf{Description} & \textbf{Value} \\
\cmidrule{1-3}
n  & Number of  rollouts & \textbf{180}\\ 
\cmidrule{1-3}
c & Exploration constant in UCT/PUCT & \textbf{0.8} \\ 
\cmidrule{1-3}
n\_updates & Number of network updates (iterations) & \textbf{50}   \\ 
\cmidrule{1-3}
n\_games & Number of self-play games per update & \textbf{1000}  \\ 
\cmidrule{1-3}
n\_envs & Number of parallel workers & \textbf{8}  \\ 
\cmidrule{1-3}
buffer\_size & Size of replay buffer & \textbf{2000} \\ 
\cmidrule{1-3}
N & Total games played ($n\_updates * n\_games$) & \textbf{50000}  \\ 
\bottomrule
\end{tabular}
\label{tab:training_param}
\end{table}

\begin{table*}
\centering
\caption{Neural network hyperparameters}

\begin{tabular}{l|m{4.25cm}|m{2.25cm}|m{2.5cm}}
\toprule
\textbf{Hyperparameter} & \multicolumn{1}{c|}{\textbf{Description}}                   & \textbf{AlphaZero Value} & \textbf{Descent Value}                      \\ 
\cmidrule(){1-4}
n\_res                  & Number of residual blocks                                   & \textbf{8} & \textbf{8}                       \\ 
\cmidrule{1-4}
res\_filters            & Number of output filters in convolutions in residual blocks & \textbf{128} & \textbf{128}                     \\ 
\cmidrule{1-4}
res\_kernel\_size       & Convolution kernel size in residual blocks                  & \textbf{3}      & \textbf{3}            \\ 
\cmidrule{1-4}
res\_activation         & Activation in residual blocks                               & \textbf{ReLU} & \textbf{ReLU}                    \\ 
\cmidrule{1-4}
policy\_filters         & Number of output filters in policy head                     & \textbf{1}          & \textbf{None}             \\ 
\cmidrule{1-4}
policy\_kernel\_size    & Convolution kenel size in policy head                       & \textbf{1}             & \textbf{None}          \\ 
\cmidrule{1-4}
policy\_activation      & Activation in the last layer of the policy head             & \textbf{Softmax}   &\textbf{None}         \\
\cmidrule{1-4}
value\_activation       & Activation function in the last layer of value head         & \textbf{Softmax}   &\textbf{Linear}                \\ 
\cmidrule{1-4}
kernel\_regularizer     & L2 regularization applied to all weights                    & \textbf{0.0001} &\textbf{0.0001}                   \\ 
\cmidrule{1-4}
policy\_loss            & Loss function used for the policy head                      & \textbf{Categorical crossentropy} &\textbf{None} \\ 
\cmidrule{1-4}
value\_loss             & Loss function used for the value head                       & \textbf{Categorical crossentropy} & \textbf{Mean squared error} \\ 
\cmidrule{1-4}
optimizer               & Training optimizer                                          & \textbf{SGD}  &\textbf{Adam}                      \\ 
\cmidrule{1-4}
lr                      & Training learning rate                                      & \textbf{0.0001}       &\textbf{0.0001}             \\

\bottomrule

\end{tabular}
\label{networkhyperparams}
\end{table*}

\subsection{Descent}
\label{subsec:Descent}

The second deep reinforcement learning algorithm is Descent \cite{CohenSolal2020LearningTP,cohen2020minimax}. Descent is a recent algorithm which has shown great success in international competitions such as the 2021 Computer Olympiad of the ICGA. Descent is not based on MCTS but on Unbounded MinMax \cite{korf_best-first_1996}.  

At the difference of Minimax, Unbounded Minimax explores the game tree in a non-homogeneous way where the exploration is a best-first approach at each iteration. In descent, the exploration is a best-first approach, that is recursively applied until the end of the game. This allows to backpropagate the values of terminal states more efficiently through the nodes chosen. Furthermore, during the exploration phase, the best move is determined by the utilization of a neural network. 

Descent architecture is the same as AlphaZero but, in AlphaZero there are two output networks (policy head and value head) and in descent, there is only one output network (value head). Furthermore, in the value head of descent, we are using a linear for the activation function as a regression for player scores.

The neural network has been trained with the value obtained from the minimax values of the trees built during the game. In addition, each state which has been explored during the game (not just for the sequences of states of the played games) is learned. As a result, all the information acquired during the search is used during the learning process. We use the same network and hyperparameters as described in the original article \cite{cohen2020minimax} and train for $120$ hours, the same duration as AlphaZero. 

In the original paper, the authors have achieved better performance that AlphaZero on multiple game and more quickly.

\section{Experimental Results}

The experiences were made on 2 NVIDIA GeForce RTX 2080 TI. Each test have been experimented on 500 games. All neural networks have been trained for $120$ hours. 

\begin{table}
\centering
\caption{Average number of point when all players are using UCT. The test has been run on $500$ game with $95\%$ confidence interval.}
\begin{tabular}{l|c|c|c}
\toprule
  & Black & White & Red \\
  \midrule

 Point & $11.5 \pm 1.0 $ & $7.1 \pm 0.9 $ & $6.3 \pm 0.9 $ \\

\bottomrule
\end{tabular}
\label{tab:all_uct}
\end{table}

As as baseline, we are using the Table \ref{tab:all_uct} . In this table, we observe the average number of points when all player are using UCT. As we are not using komi during our experimentation, it makes sense to observe that the black player has an advantage against white and red. However, as we can see, white does not have a significant advantage in comparison to the red player.

\subsection{Training of AlphaZero and Descent.}

\begin{figure*}
\centering
	\subfloat[\centering AlphaZero (Black) against UCT (White, Red).]{\includegraphics[scale=0.4]{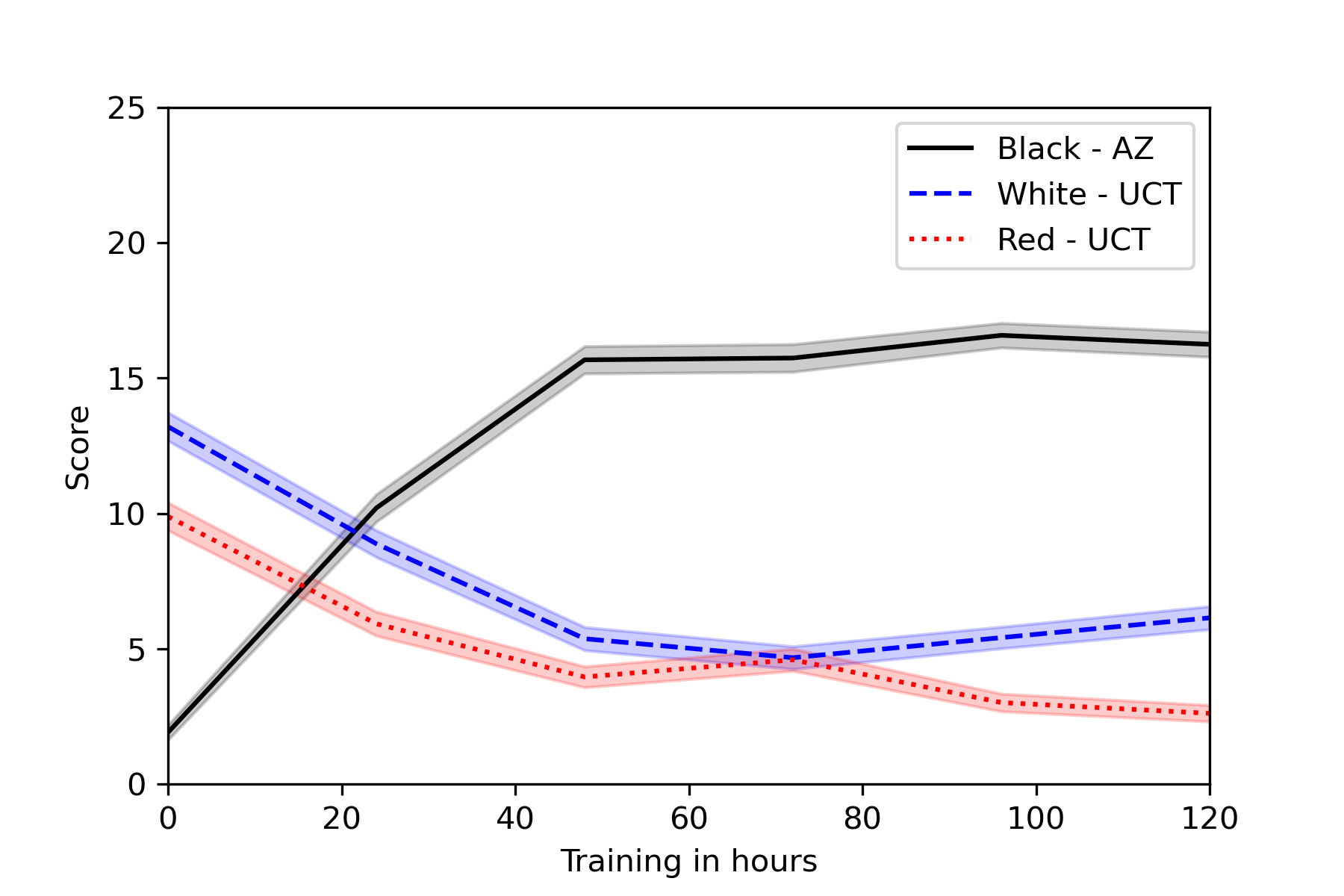}}
	\subfloat[\centering Descent (Black) against UCT (White, Red).]{\includegraphics[scale=0.4]{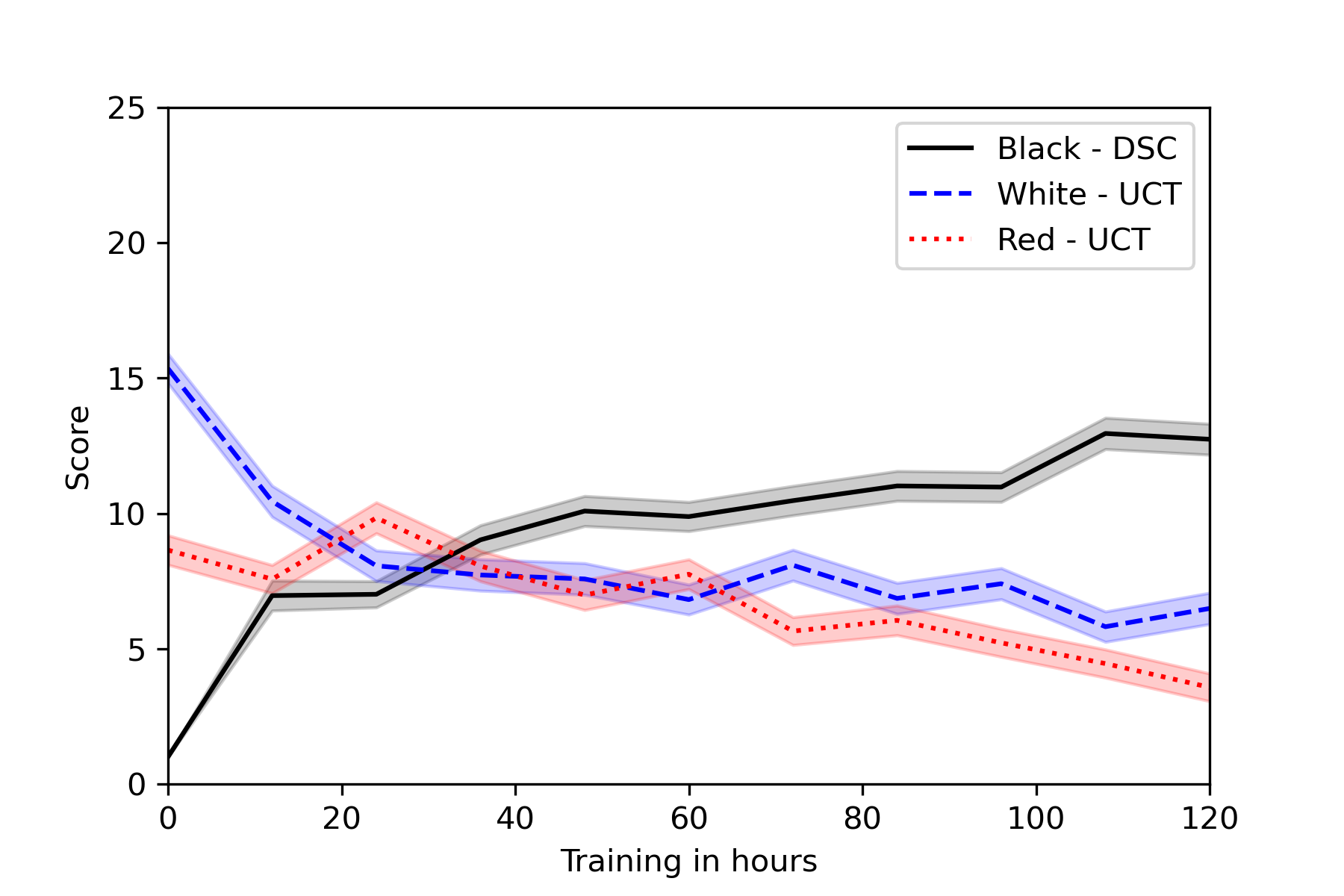}} \hfill 
	
	\subfloat[\centering AlphaZero (White) against UCT (Black, Red).]{\includegraphics[scale=0.4]{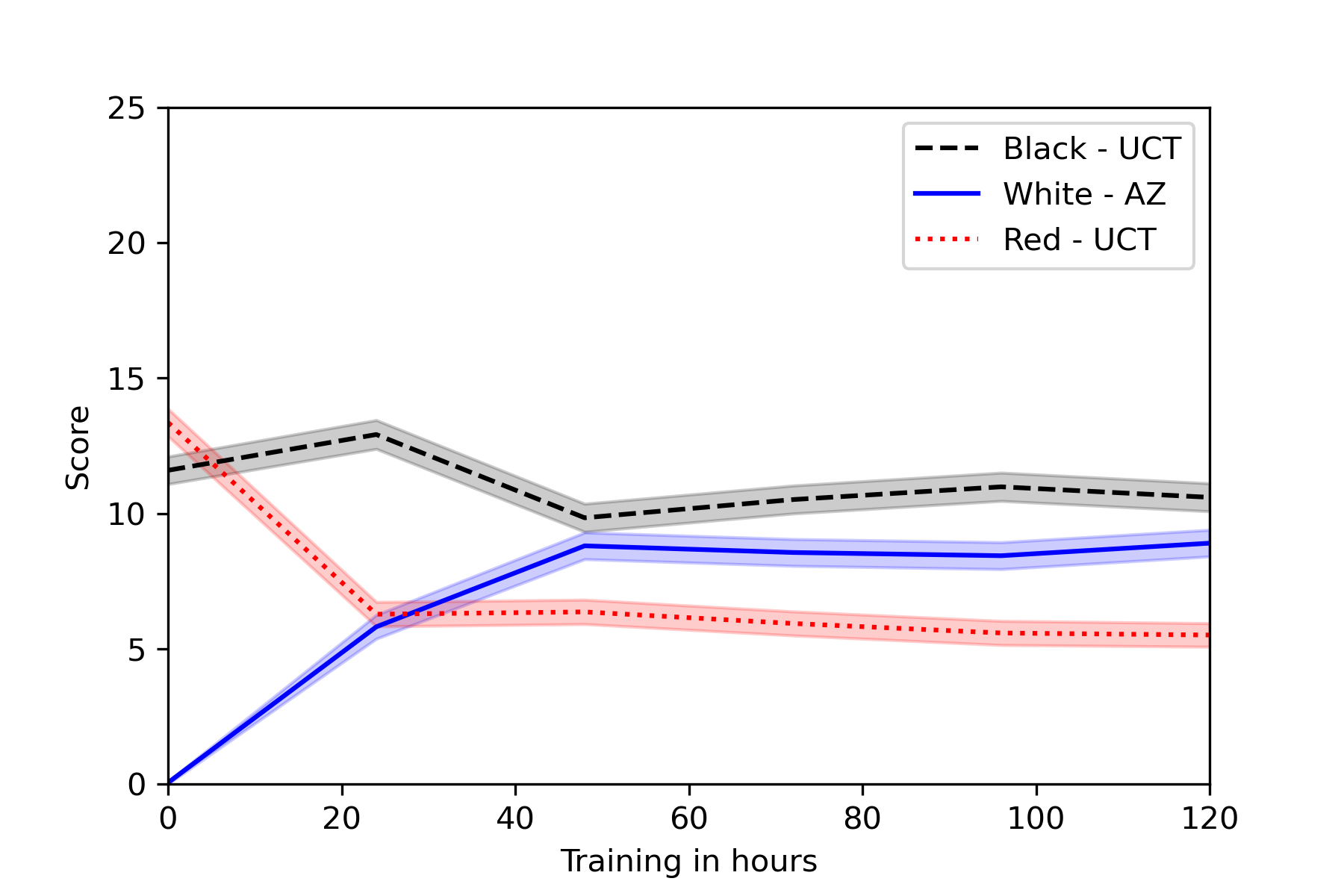}}
	\subfloat[\centering Descent (White) against UCT (Black, Red).]{\includegraphics[scale=0.4]{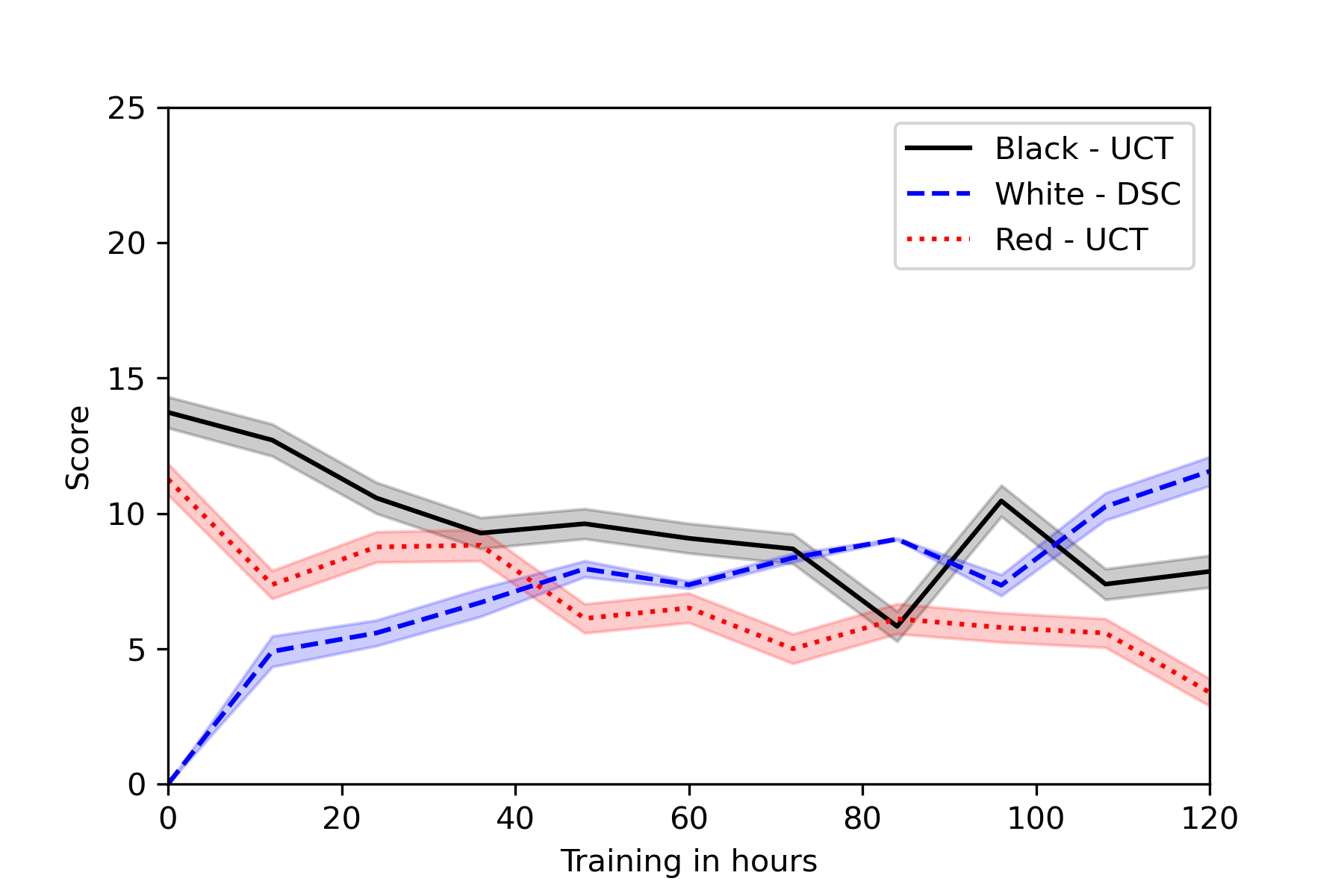}}
	
	\subfloat[\centering AlphaZero (Red) against UCT (Black, White).]{\includegraphics[scale=0.4]{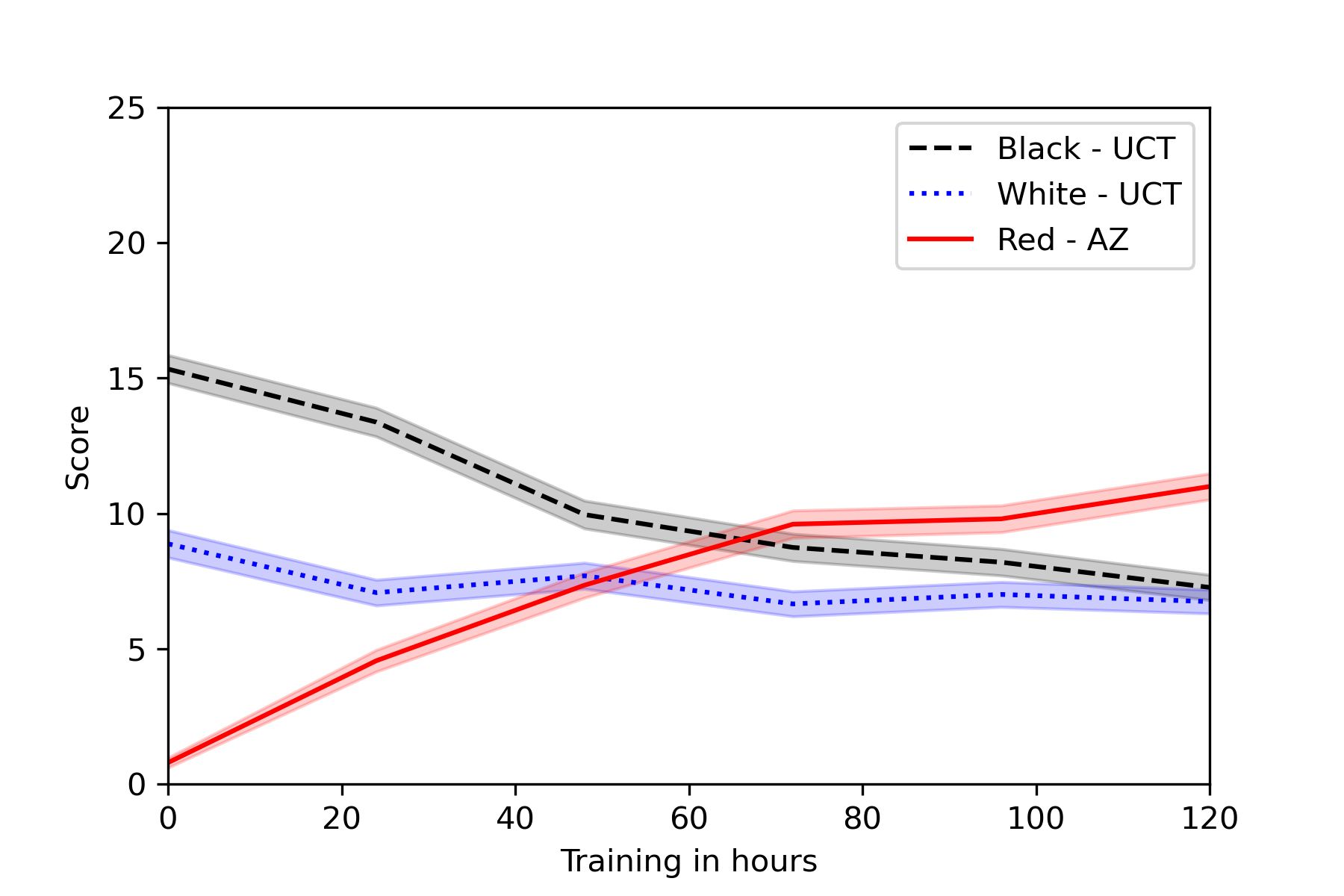}}
	\subfloat[\centering Descent (Red) against UCT (Black, White). ]{\includegraphics[scale=0.4]{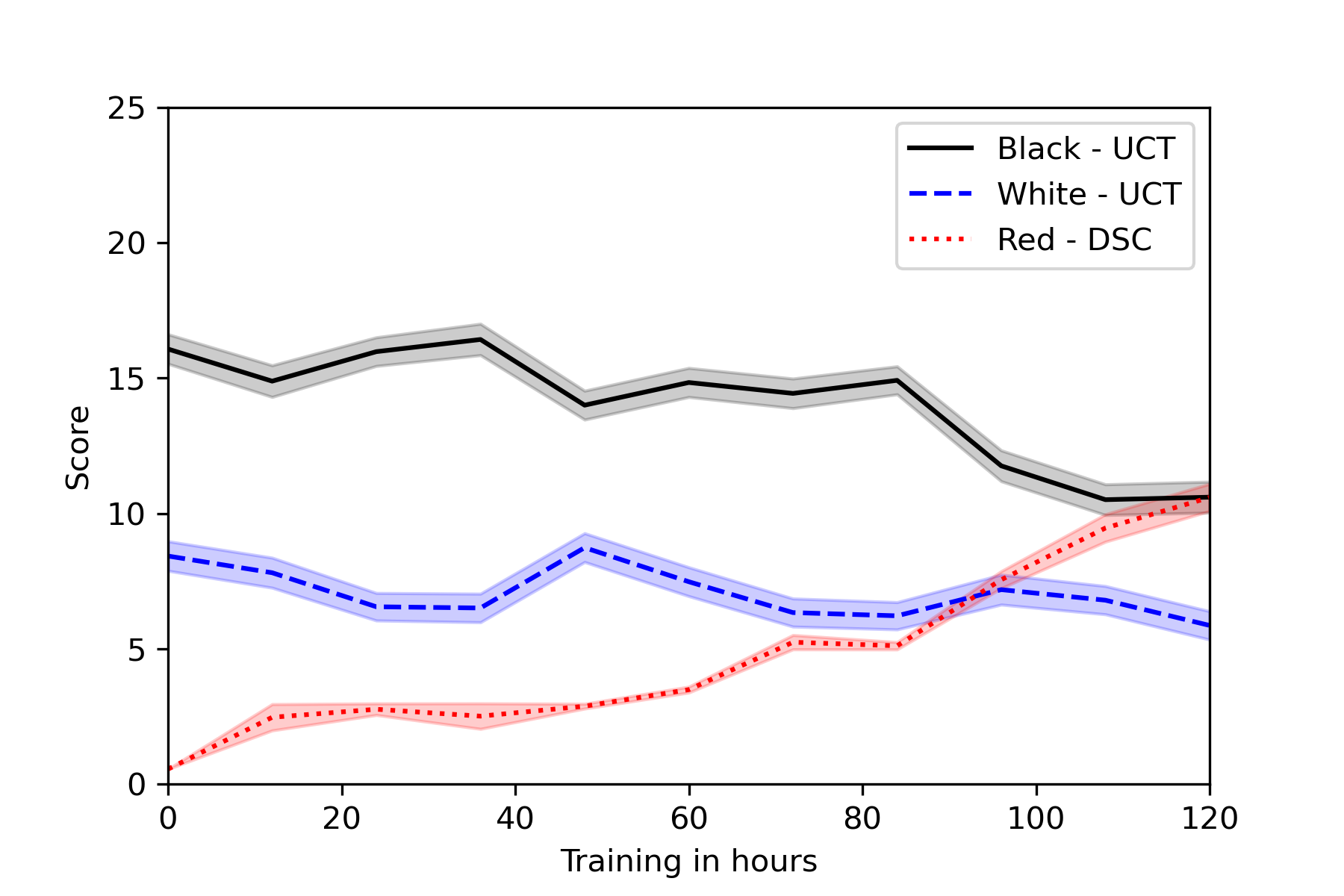}}
		
    \caption{Left/Right figure represents AlphaZero/Descent against UCT. In y-axis we observe the average points obtained and in x-axis, we observe the training in hours.}

\label{fig:az_uct}
\end{figure*}

In this subsection, we are analyzing the performance of AlphaZero and Descent against UCT for all players. In Figure \ref{fig:az_uct}, we can observe the evolution of the performance according to the training time where we tested the performance every 12 hours.

\textbf{AlphaZero}

For AlphaZero, the improvement of the performance are available in a, c and e of the Figure \ref{fig:az_uct}. As we can observer, AlphaZero improves its performance for the three different colors using the same network. At the end of the training, we observe that the average points obtained are close to $16$, $10$, $11$ for the black, white and red players respectively. 

Most of the improvement in AlphaZero has been done during the first $60$ hours of training. Even tough, we observe a stagnation after $60$ hours, the performance of AlphaZero are superior to UCT. The average score per game increased from $10$ to $16$, from $7$ to $10$ and $6$ to $11$, respectively for the black, white and red players.

In addition, when playing black and red, AlphaZero is able to outperform all other opponents, and when playing white, obtains a score almost equal to black while being at a disadvantage.
\\\\
\textbf{Descent}

For Descent, the improvement of the performance are available in b, d and f of the Figure \ref{fig:az_uct}. The figure shows that Descent also leads to performance improvement in a multiplayer setup. All the players converge to the same average score, around $11$ ($12, 12, 11$ for black, white, red respectively) and each of them having at least better or equal performance to UCT. Having different network update strategies and network architecture (Descent does not use a policy network to guide move selection), the two methods do not converge to the same performance. 


In past work of Descent, the authors had better and faster results than AlphaZero. However, as we can observe, after $120$ hours, AlphaZero has better scores for the black and red players but Descent is better for the white player. Nevertheless, we can observe that AlphaZero is not improving a lot after the $60$ hours mark whereas Descent did not start showing stagnation in its curves and this even after $120$ hours of training.

\subsection{Black against White and Red.}
AlphaZero and Descent use different planning methods. This can result in different strength and strategies while playing different positions in a game. Remember that each method use a single network for all the different positions. The UCT baseline in Table \ref{tab:all_uct} confirm that Black has an advantage when playing first, which is an expected results since it's the case in Go.
Without using komi, we will only focus on how both methods play when having this advantage.
Table \ref{tab:black_vs_other} shows the results of 500 games testing Black strengh at attacking each other weaker positions.

We notice that AlphaZero tends to be more aggressive when playing Black against itself, which is what it learn during self-play, achieving an average score of 13.3 points. The same aggression does not work effectively against Descent defenses (White and Red) since it only gets 11 points on average.
Looking at Descent scores, playing as Black against AlphaZero defenses achieves an overage score of 12.1 points, which is between both AlphaZero scores as Black. Playing against itself only show a small difference in score going to 11.9.

In both cases, Descent is stronger when playing positions at disadvantage, and does not show a bigger difference playing against AlphaZero as Black, meaning that Descent is more balanced in strengh between all the different positions.
AlphaZero on the other hand, will try to be more agressive against medium defenses (16 points against UCT and 13.3 points against itself) but this also mean that it can be slightly weaker playing White and Red and that the same strategy will not be effective against better defenses (only 11 points against Descent).

\begin{table}
\centering
\caption{Average number of point when black is against the others players for different algorithms. The test has been run on $500$ games with $95\%$ confidence interval.}
\begin{tabular}{l|c|*3{c}}

  \multicolumn{2}{c}{} & \multicolumn{3}{c}{White and Red}\\
  \cmidrule{3-5}
  \multicolumn{2}{c}{} & UCT & AlphaZero & Descent \\
  \midrule

 \multirow{2}{*}{\begin{sideways}Black\end{sideways} }
 & AlphaZero & $16.2 \pm 0.3 $  &  $13.3 \pm 0.4$ & $11 \pm 0.2$\\
 & Descent & $12.7 \pm 0.5 $ & $12.1 \pm 0.6$ & $11.9 \pm 0.3$ \\

\bottomrule
\end{tabular}
\label{tab:black_vs_other}
\end{table}

\section{Conclusion}

In this paper, we used and analyzed Deep Reinforcement Learning for one of the variants of the game of Go, the game of Multiplayer Go. We have been using AlphaZero and Descent, which have been showing great success in recent years. We demonstrate that both algorithms are applicable in Multiplayer Go and both of them are able to learn in the context of multiplayer game which is more complex than two players. 

Both of the algorithms have been able to beat or equalizes UCT in all players positions (Black, White and Red). In addition, against UCT, the two algorithms obtain very close results in a short training time and neither of the two has been able to beat the other in all cases.

In addition to this, we analyze the impact of the black player using a Deep RL algorithm against the other Deep RL algorithm for the white and red position. In this context, we show that Descent is more balanced in strength between different positions than AlphaZero which result in a better defense, but that AlphaZero can achieve better performance against medium and weaker defenses (himself or UCT) than Descent.

In future work, we expect to use Deep Reinforcement Learning on other multiplayer games, to increase the number of agents and to use it on larger boards. Furthermore, we have observed that AlphaZero stops improving after $60$ hours of training which is not the case for Descent. As a consequence, we are interested in making more and longer experiments in order to compare more accurately the two DRL algorithms.
